\documentclass[10pt,twocolumn,letterpaper]{article}
\pdfoutput=1
\usepackage{iccv}
\usepackage{times}
\usepackage{epsfig}
\usepackage{graphicx}
\usepackage{amsmath}
\usepackage{amssymb}
\usepackage[bookmarks=false]{hyperref}

\iccvfinalcopy

\begin{document}

\title{ChromaGAN: Adversarial Picture Colorization with Semantic Class Distribution}

\author{Patricia Vitoria, Lara Raad and Coloma Ballester  \\
Department of Information and Communication Technologies \\
University Pompeu Fabra, Barcelona, Spain \\
{\tt\small \{patricia.vitoria, lara.raad, coloma.ballester\}@upf.edu}
}

\maketitle

\begin{abstract}
The colorization of grayscale images is an ill-posed problem, with multiple correct solutions. In this paper, we propose an adversarial learning colorization approach coupled with semantic information. A generative network is used to infer the chromaticity of a given grayscale image conditioned to semantic clues. This network is framed in an adversarial model that learns to colorize by incorporating perceptual and semantic understanding of color and class distributions. The model is trained via a fully self-supervised strategy. Qualitative and quantitative results show the capacity of the proposed method to colorize images in a realistic way achieving state-of-the-art results.
\end{abstract}
\section{Introduction}

Colorization is the process of adding plausible color information to monochrome photographs or videos (we refer to \cite{yatziv2006fast} for an interesting historical review). Currently, digital colorization of black and white visual data is a crucial task in areas so diverse as advertising and film industries, photography technologies or artist assistance. 
Although important progress has been achieved in this field, automatic image colorization still remains a challenge. 

\begin{figure}[t]
\begin{center}
   \includegraphics[width=0.3\linewidth]{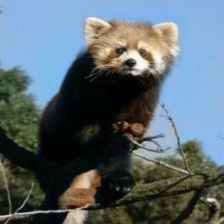}
      \includegraphics[width=0.3\linewidth]{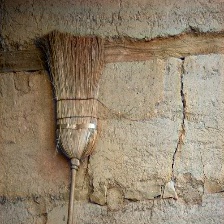}
    \includegraphics[width=0.3\linewidth]{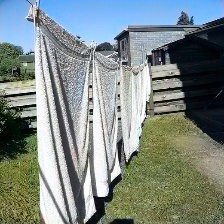}
    \includegraphics[width=0.3\linewidth]{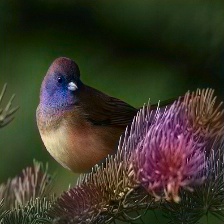}
       \includegraphics[width=0.3\linewidth]{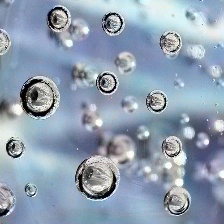}
      \includegraphics[width=0.3\linewidth]{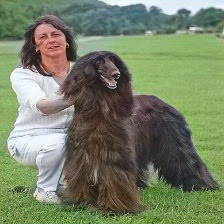}
    \includegraphics[width=0.3\linewidth]{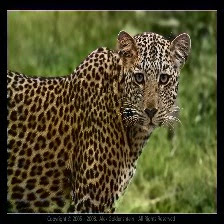}
    \includegraphics[width=0.3\linewidth]{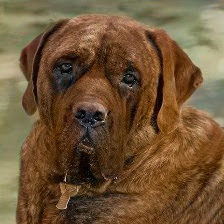}
    \includegraphics[width=0.3\linewidth]{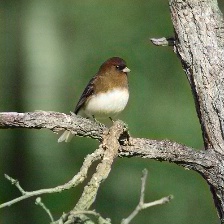}
\end{center}
   \caption{ChromaGAN is able to colorize a grayscale image from the semantic understanding of the captured scene.}
\label{fig:SomeResults}
\end{figure}

Colorization is a highly undetermined problem, requiring mapping a real-valued grayscale image to a three-dimensional color-valued one, that has not a unique solution.
Before the emergence of deep learning techniques, the most effective methods relied on human intervention, usually through either user-provided color scribbles or a color reference image.  Recently, convolutional neural network strategies have benefit from the huge amount of publicly available color images in order to automatically learn what colors naturally correspond to the real objects and its parts.
In this paper, we propose a fully automatic end-to-end adversarial approach called ChromaGAN. It combines the strength of generative adversarial networks (GANs) with semantic class distribution learning. 
As a result, ChromaGAN is able to perceptually colorize a grayscale image from the semantic understanding of the captured scene. To illustrate this, Fig.~\ref{fig:SomeResults} shows vibrant and diverse colorizations frequently achieved. 
Moreover, ChromaGAN shows variability by colorizing differently some objects belonging to the same category that may have several real colors, as for example, the birds in Fig.~\ref{fig:SomeResults}.
The user-based perceptual ablation study 
show that the effect of the generative adversarial learning is key to obtain those vivid colorizations. 

The contributions of this work include: 

    \noindent
    $\bullet$ An adversarial learning approach coupled with semantic information leading to a three term loss combining color and perceptual information with semantic class distribution.\\ 
    \noindent
    $\bullet$ An unsupervised semantic class distribution learning.\\
    \noindent
    $\bullet$ A perceptual study showing that semantic clues coupled to an adversarial approach yields high quality results.

The outline of this paper is as follows. Section~\ref{sec:relwork} reviews the related work. In Section~\ref{sec:method} the proposed model, architecture and algorithm are detailed. Section~\ref{sec:results} presents qualitative and quantitative results. Finally, the paper is concluded in Section~\ref{sec:conclusion}. The code is available at https://github.com/pvitoria/ChromaGAN. 

\section{Related Work}\label{sec:relwork}

In the past two decades, several colorization techniques have been proposed. They can be classified in three classes: scribble-based, exemplar-based and deep learning-based methods. The first two classes depend on human intervention. The third one is based on 
learning leveraging the possibility of easily creating training data from any color image.

\textbf{Scribble-based methods.} The user provides local hints, as for instance color scribbles, which are then propagated to the whole image. These methods were initiated with the work of Levin \etal~\cite{levin2004colorization}. They assume that spatial neighboring pixels having similar intensities should have similar colors. They formalize {{this premise}} optimizing a quadratic cost function constrained to the values given by the scribbles.
Several improvements were proposed in the literature. Huang \etal~\cite{huang2005adaptive} improve the bleeding artifact using the grayscale image edge information. Yatziv \etal~\cite{yatziv2006fast} propose a luminance-weighted chrominance blending to relax the dependency of the position of the scribbles. Then, Luan \etal~\cite{luan2007natural} use the scribbles to segment the grayscale image and thus better propagate the colors. {{These methods}} suffer from requiring large amounts of user inputs in particular when dealing with complex textures. Moreover, choosing the correct color palette is not a trivial task.

\textbf{Exemplar-based methods.} They transfer the colors of a reference image to a grayscale one. Welsh \etal ~\cite{welsh2002transferring}, propose to match the luminance values and texture information between images. This approach lacks of spatial coherency which yields unsatisfactory results. To overcome this, several improvements have been proposed. Ironi \etal~\cite{ironi2005colorization} transfer some color values from a segmented source image which are then used as scribbles in~\cite{levin2004colorization}. Tai \etal~\cite{tai2005local} construct a probabilistic segmentation of both images to transfer color between any two regions having similar statistics. Charpiat \etal~\cite{charpiat2008automatic} deal with the multimodality of the colorization problem estimating for each pixel the conditional probability of colors. Chia \etal~\cite{chia2011semantic} use the semantic information of the grayscale image. Gupta \etal~\cite{gupta2012image} transfer colors based on the features of the superpixel representation of both images. Bugeau \etal~\cite{bugeau2014variational} colorize an image by solving a variational model which allows to select the best color candidate, from a previous selection of color values, while adding some regularization in the colorization. Although this type of methods reduce significantly the user inputs, they are still highly dependent on the reference image which must be similar to the grayscale image.

\textbf{Deep learning methods.} Recently, different approaches have been proposed to leverage the huge amount of grayscale/color image pairs. Cheng \etal~\cite{cheng2015deep} first proposed a fully-automatic colorization method formulated as a least square minimization problem solved with deep neural networks. A semantic feature descriptor is proposed and given as an input to the network. In~\cite{deshpande2015learning}, a supervised learning method is proposed through a 
linear parametric model and a variational autoencoder which is computed by quadratic regression on a dataset of color images. 
These approaches are improved by the use of CNNs and large-scale datasets. For instance, Iizuka \etal~\cite{iizuka2016let} extract 
local and global features to predict the colorization. 
The network is trained jointly for classification and colorization in a labeled dataset.
Zhang \etal~\cite{zhang2016colorful} learn the color distribution of every pixel.
The network is trained with a multinomial cross entropy loss with rebalanced rare classes allowing unusual colors to appear. Mouzon \etal~\cite{mouzon2019joint} couple the resulting distribution in Zhang \etal~\cite{zhang2016colorful} with the variational approach proposed in~\cite{pierre2015luminance}. This allows to select for each pixel a color candidate from the pixel color distributions while regularizing the result. Also, it avoids the halo artifacts noticed in~\cite{zhang2016colorful}. Larsson \etal~\cite{larsson2016learning} train a deep CNN to learn per-pixel color histograms. They use a VGG network to interpret the semantic composition of the scene and the localization of objects to then predict the color histograms of every pixel. The network is trained with the KL divergence. 

Other CNN based approaches are combined with user interactions. For instance, Zhang \etal~\cite{zhang2017real} propose to train a deep network given the grayscale version and a set of sparse user inputs. This allows the user to have more than one plausible solution. Also, He \etal~\cite{he2018deep} propose an exemplar-based colorization method using a deep learning approach. The colorization network jointly learns faithful local colorization to a meaningful reference and plausible color prediction when a reliable reference is unavailable. These hybrid methods yield results containing rare colors. Recently, Yoo \etal in~\cite{yoo2019coloring} capture rare color instances without human interaction using a memory network together with a triplet loss without the need of labels.

Some methods use GANs \cite{GAN} to colorize grayscale images. Their ability in learning probability distributions over high-dimensional spaces of data such as color images has found widespread use for many tasks 
(\eg, \cite{chan2018everybody,isola2017image,karras2018style,lin2017adversarial,pumarola2018ganimation,vitoria2018semantic,zhu2017unpaired}).  
Isola \etal~\cite{isola2017image} propose to use conditional GANs to map an input image to an output image using a U-Net based generator. They train their network by combining the $L_1$-loss with an adapted GAN loss. An extension is proposed in~\cite{nazeri2018image} where the authors claim to generalize~\cite{isola2017image} to high resolution images, speed up and stabilize the training. Cao \etal~\cite{cao2017unsupervised} also use conditional GANs and obtain diverse possible colorizations by sampling several times the input noise. 
It is incorporated in multiple layers in their 
architecture, which consists of a fully convolutional non-stride network. Notice that none of these GANs based methods use additional information such as classification while our method incorporates 
the distribution of semantic classes in an adversarial approach coupled with color regression.

\begin{figure*}[t]
\begin{center}
   \includegraphics[width=\linewidth]{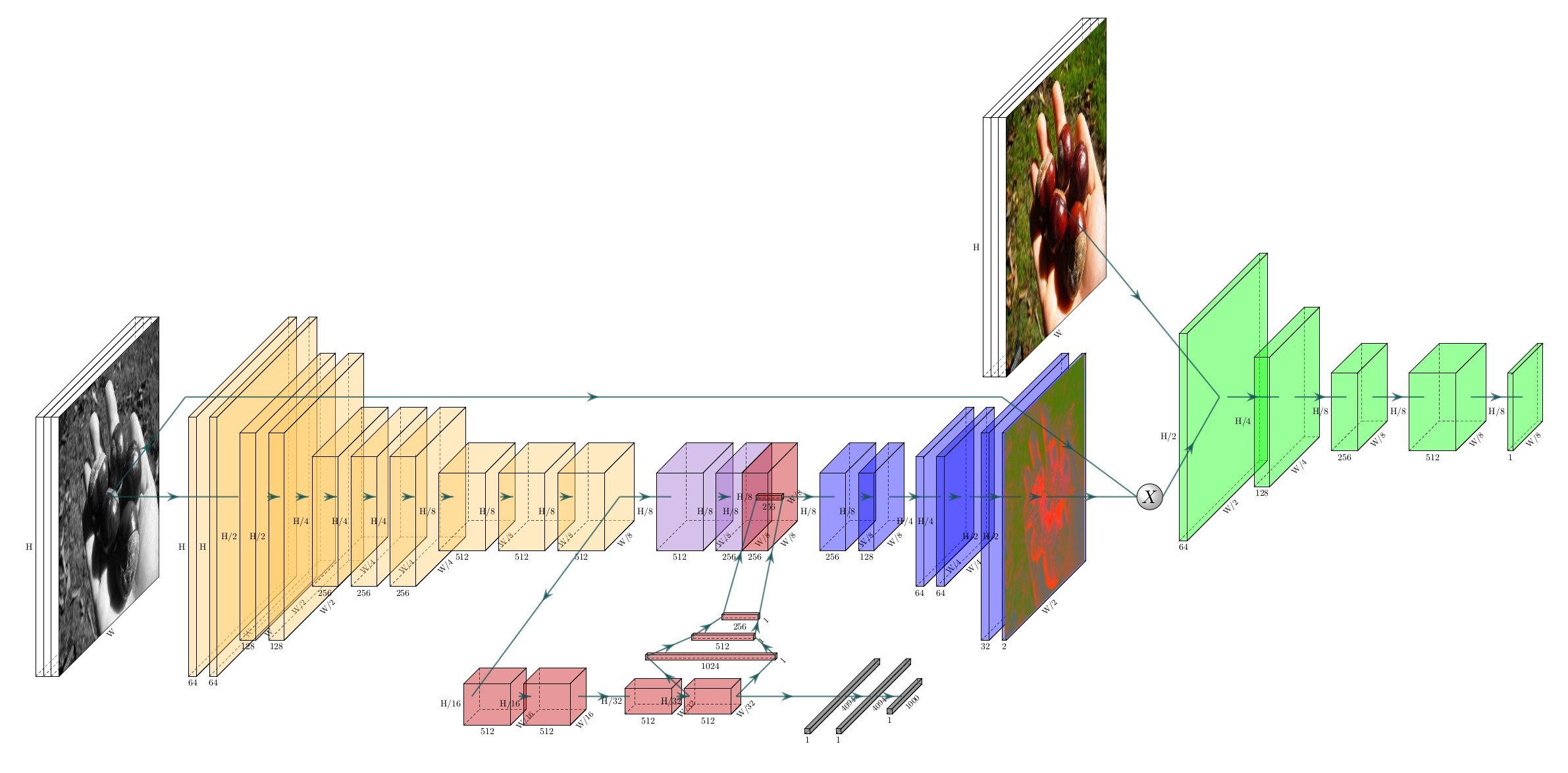}
\end{center}
   \caption{Overview of our model, ChromaGAN, able to automatically colorize grayscale images. It combines a Discriminator network, $D_w$ (in green), and a Generator network, ${\cal{G}}_\theta$. ${\cal{G}}_\theta$ consists of two subnetworks:  ${\cal{G}}^1_{\theta_1}$ (yellow, purple, red and blue layers) that outputs the chrominance information $(a,b)={\cal{G}}^1_{\theta_1}(L)$, and ${\cal{G}}^2_{\theta_2}$ (yellow, red and gray layers) which outputs the class distribution vector, $y={\cal{G}}^2_{\theta_2}(L)$.
   }
\label{fig:model}
\end{figure*}

\section{Proposed Approach}\label{sec:method}

Given a grayscale input image $L$, we learn a mapping ${\cal{G}}: L\longrightarrow (a,b)$ such that $I=(L,a,b)$ is a plausible color image and $a$ and $b$ are chrominance channel images in the CIE $Lab$ color space. A plausible color image is one having geometric, perceptual and semantic photo-realism.

In this paper, the mapping ${\cal{G}}$ is learnt by means of an adversarial learning strategy.  The colorization is produced through a generator $-$equivalent to ${\cal{G}}$ above$-$ 
that predicts the chrominance channels $(a,b)$. In parallel, a discriminator evaluates how realistic is the proposed colorization $I=(L,a,b)$ of $L$.
To this aim, we propose in Section~\ref{ssec:energy} a new adversarial energy that learns the parameters $\theta$ and $w$ of the generator ${\cal{G}}_\theta$ and the discriminator $D_w$, respectively. This is done training end-to-end the proposed network in a self-supervised manner by using a dataset 
of real color images.
In particular, given a color image $I_r=(L,a_r,b_r)$ in the CIE $Lab$ color space, we train our model to learn the color information by detaching the training data in two parts: The grayscale $L$ and the chrominance channels  $(a_r,b_r)$. 

For the sake of clarity and by a slight abuse of notation, we shall write ${\cal{G}_\theta}$ and $D_w$ instead of $\theta$ and $w$, respectively. Moreover, our generator ${\cal{G}_\theta}$ will not only learn to generate color but also a class distribution vector, denoted by $y\in\mathbb{R}^m$, where $m$ is the fixed number of classes. This provides information about the probability distribution of the semantic content and objects present in the image. 
The authors of ~\cite{iizuka2016let} also incorporate semantic vectors but their proposal needs class labels while we learn the semantic image distribution contained in $L$ that boosts its colorization without any need of a labeled dataset. For that, our generator model combines two different modules. 
We denote it by ${\cal{G}}_\theta=({\cal{G}}^1_{\theta_1},{\cal{G}}^2_{\theta_2})$, where $\theta=(\theta_1,\theta_2)$ stand for all the generator parameters, ${\cal{G}}^1_{\theta_1}: L\longrightarrow (a,b)$, and  ${\cal{G}}^2_{\theta_2}: L\longrightarrow y$.
An overview of the model architecture can be seen in Fig.~\ref{fig:model} and will be described in Section~\ref{ssec:archit}. In the next Section~\ref{ssec:energy} the proposed adversarial loss is stated.

\subsection{The Objective Function}\label{ssec:energy}
Our objective loss is defined by
\begin{equation}\label{eq:loss}
    {\cal{L}}({\cal{G}}_\theta,D_w) = 
    {\cal{L}}_{\text{e}}({\cal{G}}^1_{\theta_1}) 
    +\lambda_{\text{g}} {\cal{L}}_{\text{g}}({\cal{G}}^1_{\theta_1},D_w)
    +\lambda_{\text{s}} {\cal{L}}_{\text{s}}({\cal{G}}^2_{\theta_2}).
\end{equation}
The first term denotes the \textit{color error loss}
\begin{equation}\label{eq:errorloss}
    {\cal{L}}_{\text{e}}({\cal{G}}^1_{\theta_1})
    ={\mathbb{E}}_{(L,a_r,b_r)\sim {\mathbb{P}}_r}\left[\| {\cal{G}}^1_{\theta_1} (L) - (a_r,b_r))\|_{2}^2\right] 
\end{equation}
where ${\mathbb{P}}_r$ stands for the distribution of real color images and $\|\cdot\|_{2}$ for the  Euclidean norm. Notice that Euclidean distance in the $Lab$ color space is more adapted to perceptual color differences.
Then,
\begin{equation}\label{eq:classloss}
    {\cal{L}}_{\text{s}}({\cal{G}}^2_{\theta_2})=
    {\mathbb{E}}_{L\sim {\mathbb{P}}_{rg}}\left[\text{KL}\left(y_{\text{v}}\ \| \, {\cal{G}}^2_{\theta_2}(L)\right)\right]
\end{equation}
denotes the \textit{class distribution loss}, where ${\mathbb{P}}_{rg}$ denotes the distribution of grayscale input images, and $y_{\text{v}}\in\mathbb{R}^m$ the output distribution vector of a pre-trained VGG-16 model applied to the grayscale image \cite{simonyan2014very} (details are given below).  $\text{KL}(\cdot\|\cdot)$ stands for the Kullback-Leibler divergence. 

Finally, ${\cal{L}}_{\text{g}}$ denotes the \textit{WGAN loss} which consists of an adversarial Wasserstein GAN loss~\cite{WGAN}. Let us first remark that leveraging the WGAN instead of other GAN losses favours nice properties such as avoiding vanishing gradients and mode collapse, and achieves more stable training. To compute it, we use the Kantorovich-Rubinstein duality~\cite{kantoro,villani2008optimal}. As in \cite{WGAN-GP}, we also include a gradient penalty term constraining the $L_2$ norm of the gradient of the discriminator with respect to its input and, thus, imposing that $D_w\in{\cal D}$,  where ${\cal{D}}$ denotes the set of 1-Lipschitz functions. To sum up, our WGAN loss is defined by
\begin{equation}\label{eq:wganloss}
\begin{split}
    {\cal{L}}_{\text{g}}({\cal{G}}^1_{\theta_1},D_w) & =  {\mathbb{E}}_{\tilde{I}\sim {\mathbb{P}}_r}\left[ D_w(\tilde{I})\right] \\ 
    & - {\mathbb{E}}_{(a,b)\sim {\mathbb{P}}_{{\cal{G}}^1_{\theta_1}}}\left[ D_w(L,a,b)\right]\\
    & - \mathbb{E}_{\hat{I}\sim {\mathbb{P}}_{\hat{I}}}[(\|\nabla_{\hat{I}}D_w(\hat{I})\|_2-1)^2]. 
\end{split}
\end{equation}
where ${\mathbb{P}_{{\cal{G}}^1_{\theta_1}}}$ is the model distribution of ${\cal{G}}^1_{\theta_1}(L)$, with $L\sim {\mathbb{P}}_{rg}$. As in ~\cite{WGAN-GP}, ${\mathbb{P}}_{\hat{I}}$ is implicitly defined sampling uniformly along straight lines between pairs of point sampled from the data distribution $\mathbb{P}_r$ and the generator distribution ${\mathbb{P}_{{\cal{G}}^1_{\theta_1}}}$. 
The minus before the gradient penalty term in \eqref{eq:wganloss} corresponds to the fact that, in practice, when optimizing with respect to the discriminator parameters, our algorithm minimizes the negative of the loss instead of maximizing it. 

From the previous loss~\eqref{eq:loss}, we compute the weights of ${\cal{G}}_\theta,D_w$ by solving the following min-max problem 
\begin{equation}\label{eq:gan}
\min_{{\cal{G}}_\theta} \max_{D_w\in{\cal D}} \, {\cal{L}}({\cal{G}}_\theta,D_w),
\end{equation}
The hyperparameters $\lambda_{\text{g}}$ and $\lambda_{\text{s}}$ are fixed and set to $0.1$ and $0.003$, respectively. Let us comment more in detail the benefits of each of the elements of our approach.

\textbf{The adversarial strategy and the GAN loss ${\cal{L}}_{\text{g}}$.} 
The min-max problem~\eqref{eq:gan} follows the usual generative adversarial game between the generator and the discriminator networks. 
The goal 
is to learn the parameters of the generator so that the probability distribution of the generated data approaches the one of the real data, while the discriminator aims to distinguish between them. 
The initial GAN proposals optimize the Jensen-Shannon divergence that can be non-continuous with respect to the generator parameters. Besides, the WGAN~\cite{WGAN,arjovsky2017wasserstein} minimizes an approximation of the Earth-Mover distance 
between two probability distributions. 
It is known to be a powerful tool to compare 
distributions with non-overlapping supports, in contrast to the KL and the Jensen-Shannon divergences which produce the vanishing gradients problem.
Also, the WGAN alleviates the mode collapse problem which is interesting when aiming to be able to capture multiple possible colorizations.

{The proposed ${\cal{L}}_{\text{g}}$ loss favours perceptually real results.} As the experiments in Section~\ref{sec:results} show and has been also noticed by some  authors~\cite{isola2017image}, the adversarial GAN model produces sharp and colorful images favouring the emergence of a perceptually real palette of colors instead of ochreish outputs produced by colorization using only terms such as the $L_2$ or $L_1$ color error loss. {It is in agreement to the analysis in~\cite{blau2018perception}. The authors define the perceptual quality of an image as the degree to which it looks like a natural image  and argue that it can be defined as the best possible probability of success in real-vs-fake user studies. They show that perceptual quality is proportional to the distance between the distribution of the generated images and the distribution of natural images, which, at the optimum generator and discriminator, corresponds to the ${\cal{L}}_{\text{g}}$ value.}

\textbf{Color Error Loss.}
In some colorization methods~\cite{larsson2016learning,zhang2016colorful} the authors propose to learn a per-pixel color probability distribution allowing them to use different classification losses. Instead, we chose to learn two chrominance values $(a,b)$ per-pixel using the $L_2$ norm. As mentioned, only using this type of loss yields ochreish outputs. However, in our case the use of the perceptual GAN-based loss relaxes this effect making it sufficient to obtain notable results. %

\textbf{Class Distribution Loss.}
The KL-based loss  ${\cal{L}}_{\text{s}}({\cal{G}}^2_{\theta_2})$ \eqref{eq:classloss} compares the generated density distribution vector $y={\cal{G}}^2_{\theta_2}(L)$ to the ground truth distribution $y_{\text{v}}\in\mathbb{R}^m$. The latter is computed using the VGG-16~\cite{simonyan2014very} pre-trained on ImageNet dataset~\cite{imagenet_cvpr09}. The VGG-16 model was trained on color images; in order to use it without any further training, we re-shape the grayscale image as $(L,L,L)$. 
The class distribution loss adds semantic interpretation of the scene. The effect of this term is analyzed in Section~\ref{sec:results}.

\subsection{Detailed Model Architecture}\label{ssec:archit}
The proposed GAN architecture is conditioned by the grayscale image $L$ through the proposed loss~\eqref{eq:loss}. 
It contains three distinct parts. The first two, belonging to the generator, focus on geometrically and semantically generating a color image information $(a,b)$ and classifying its semantic content. The third one belongs to the discriminator network learning to distinguish between real and fake data. 

An overview of the model is shown in Fig.~\ref{fig:model}. In the remaining of the section we will describe the full architecture.
More details are available in the supplementary material.


\textbf{Generator Architecture.}
The generator ${\cal{G}}_\theta$ is made of two subnetworks (denoted by ${\cal{G}}^1_{\theta_1}$ and ${\cal{G}}^2_{\theta_2}$) divided in three stages with some shared modules between them. Both of them will take as input a grayscale image of size $H\times W$. The subnetwork ${\cal{G}}^1_{\theta_1}$ outputs the chrominance information, $(a,b)={\cal{G}}^1_{\theta_1}(L)$, and the subnetwork ${\cal{G}}^2_{\theta_2}$ outputs the computed class distribution vector, $y={\cal{G}}^2_{\theta_2}(L)$. Both subnetworks are jointly trained with a single step backpropagation.

The first stage (displayed in yellow in Fig.  \ref{fig:model}) is shared between both subnetworks. It has the same structure as the VGG-16 without the three last fully-connected layers at the top of the network. We initialize them with pre-trained VGG-16 weights which are not frozen during training.

From this first stage on, both subnetworks, ${\cal{G}}^1_{\theta_1}$ and ${\cal{G}}^1_{\theta_2}$, split into two  tracks. The first one (in purple in Fig. \ref{fig:model}) process the data by using two modules of the form Conv-BatchNorm-ReLu. 
The second track (in red), 
first processes the data by using four modules of the form Conv-BatchNorm-ReLu, followed by three fully connected layers (in red). 
This second path (in gray) 
outputs ${\cal{G}}^2_{\theta_2}$ providing the class distribution vector. To generate the probability distribution $y$ of the $m$ semantic classes, we apply a softmax function. Notice that ${\cal{G}}^2_{\theta_2}(L)$ is a classification network and is initialized with pre-trained weights. But, as part of this subnetwork is shared with ${\cal{G}}^1_{\theta_1}$, once the network is trained, it has learnt to give a class distribution close to the output of the VGG-16 and to generate useful information to help the colorization process. This could be understood as fine tuning the network  to learn to perform two tasks at once. 

In the third stage both branches are fused (in red and purple in Fig. \ref{fig:model}) by concatenating the output features. Later on, $(a,b)$ will be predicted by processing the information through six modules of the form Convolution-ReLu with two up-sampling layers in between. Note that while performing back propagation with respect to the class distribution loss, only the second subnetwork ${\cal{G}}^2_{\theta_2}$ will be affected. For the color error loss, the entire network will be affected. 


\textbf{Discriminator Architecture.}
The discriminator network $D_w$ is based on the Markovian discriminator architecture (PatchGAN~\cite{isola2017image}). The PatchGAN discriminator keeps track of the high-frequency structures of the generated image compensating the fact that the $L_2$ loss ${\cal{L}}_\text{e}({\cal{G}}^1_{\theta_1})$ fails in capturing high-frequency structures but succeeds in capturing low-level ones. In order to model the high-frequencies, the PatchGAN discriminator focuses on local patches. {Thus, rather than giving a single output for the full image, it classifies each patch as real or fake.}  We follow the architecture defined in~\cite{isola2017image} where the input and output are of size $H\times W$ and $H/8\times W/8$, respectively.

\begin{figure*}[]
    \centering
    \begingroup
\setlength{\tabcolsep}{1pt} 
    \begin{tabular}{ccccccc}
    \textbf{GT}  & \textbf{ChromaGAN}  & \textbf{w/o Class} & \textbf{Chroma Net.} & \textbf{Iizuka  \cite{iizuka2016let}} & \textbf{Larsson  \cite{larsson2016learning}} & \textbf{Zhang  \cite{zhang2016colorful}} \\
    
    \includegraphics[width=0.13\linewidth]{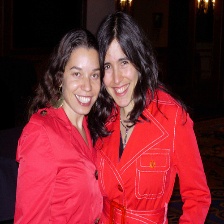} &
    \includegraphics[width=0.13\linewidth]{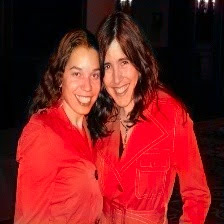} &
        \includegraphics[width=0.13\linewidth]{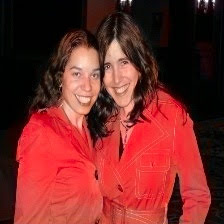} &
   \includegraphics[width=0.13\linewidth]{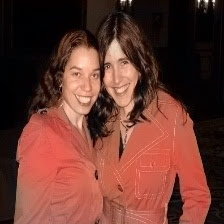} &
      \includegraphics[width=0.13\linewidth]{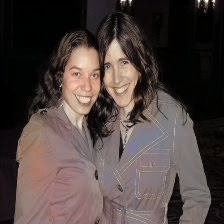}  &
     \includegraphics[width=0.13\linewidth]{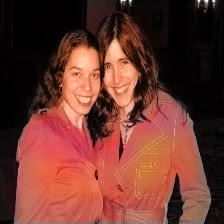} &
 \includegraphics[width=0.13\linewidth]{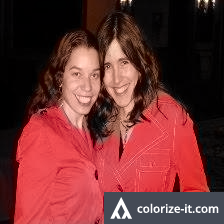} \\

       \includegraphics[width=0.13\linewidth]{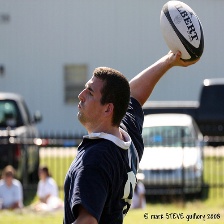} &
   \includegraphics[width=0.13\linewidth]{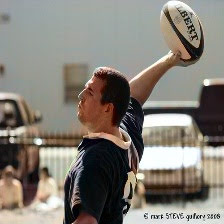} &
   \includegraphics[width=0.13\linewidth]{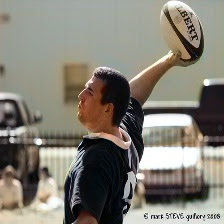} &
   \includegraphics[width=0.13\linewidth]{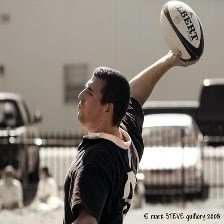} &
      \includegraphics[width=0.13\linewidth]{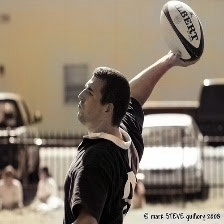} &
\includegraphics[width=0.13\linewidth]{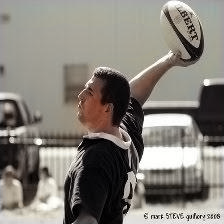} &
 \includegraphics[width=0.13\linewidth]{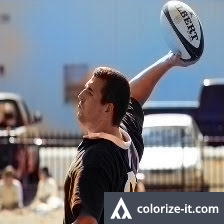}
 
  \\

   \includegraphics[width=0.13\linewidth]{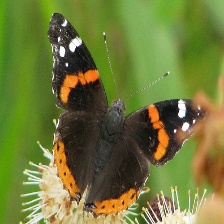} &
    \includegraphics[width=0.13\linewidth]{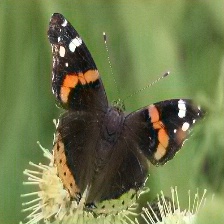} &
        \includegraphics[width=0.13\linewidth]{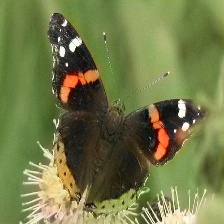} &
   \includegraphics[width=0.13\linewidth]{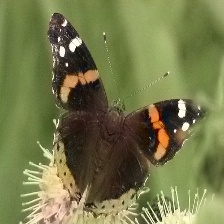} &
      \includegraphics[width=0.13\linewidth]{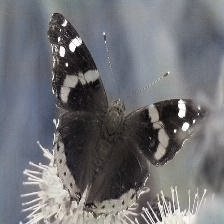} &
     \includegraphics[width=0.13\linewidth]{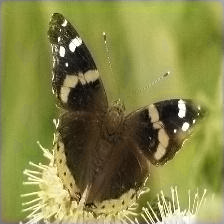} &
 \includegraphics[width=0.13\linewidth]{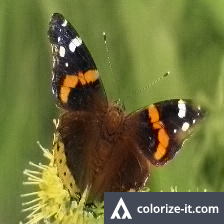} \\

    \includegraphics[width=0.13\linewidth]{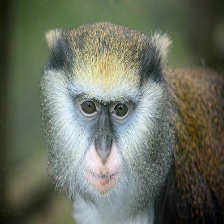}&
    \includegraphics[width=0.13\linewidth]{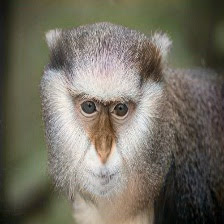} &
   \includegraphics[width=0.13\linewidth]{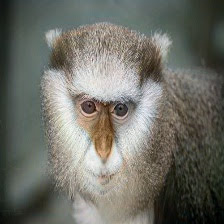} &
   \includegraphics[width=0.13\linewidth]{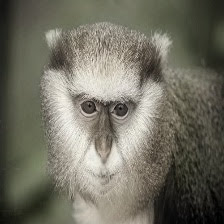} &
      \includegraphics[width=0.13\linewidth]{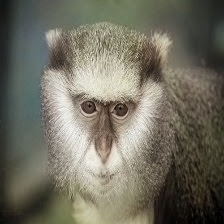} &
            \includegraphics[width=0.13\linewidth]{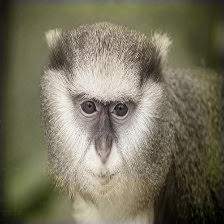} &
 \includegraphics[width=0.13\linewidth]{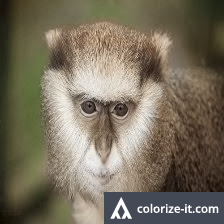} \\
 
  \includegraphics[width=0.13\linewidth]{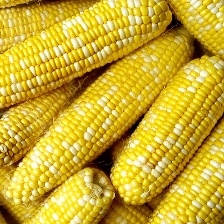} &
    \includegraphics[width=0.13\linewidth]{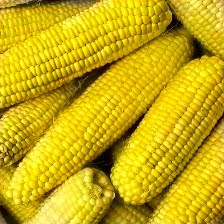} &
   \includegraphics[width=0.13\linewidth]{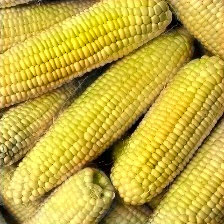} &
   \includegraphics[width=0.13\linewidth]{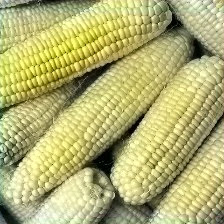} &
      \includegraphics[width=0.13\linewidth]{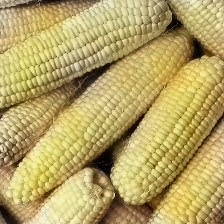} &
         \includegraphics[width=0.13\linewidth]{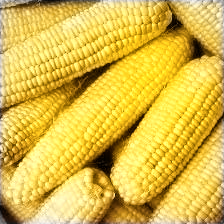} &
 \includegraphics[width=0.13\linewidth]{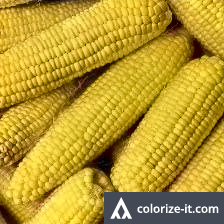}  \\
 \includegraphics[width=0.13\linewidth]{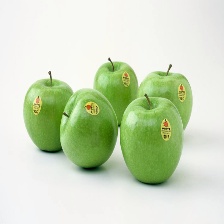} &
    \includegraphics[width=0.13\linewidth]{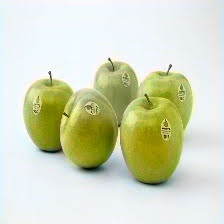} &
   \includegraphics[width=0.13\linewidth]{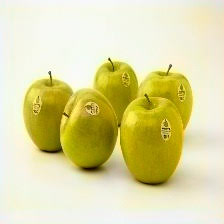} &
   \includegraphics[width=0.13\linewidth]{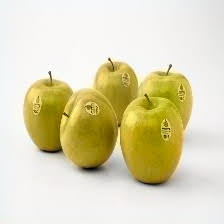} &
      \includegraphics[width=0.13\linewidth]{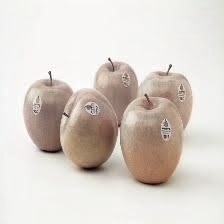} &
   \includegraphics[width=0.13\linewidth]{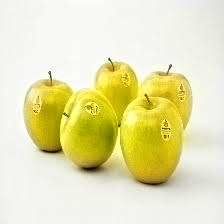} &
 \includegraphics[width=0.13\linewidth]{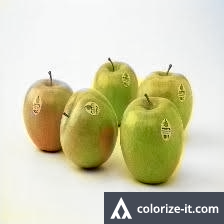} \\

   \includegraphics[width=0.13\linewidth]{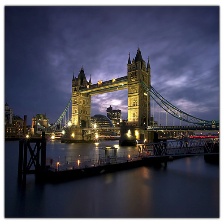} &
    \includegraphics[width=0.13\linewidth]{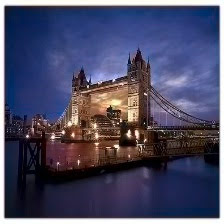} &
        \includegraphics[width=0.13\linewidth]{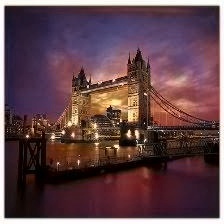} &
   \includegraphics[width=0.13\linewidth]{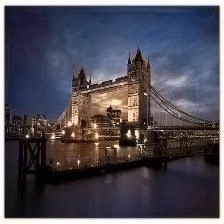} &
      \includegraphics[width=0.13\linewidth]{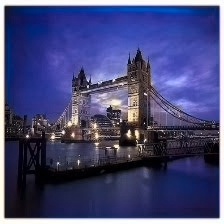}     &
     \includegraphics[width=0.13\linewidth]{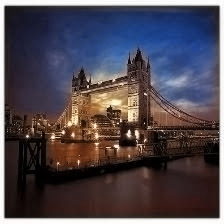} &
 \includegraphics[width=0.13\linewidth]{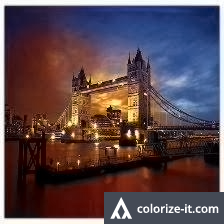} \\
 
   \includegraphics[width=0.13\linewidth]{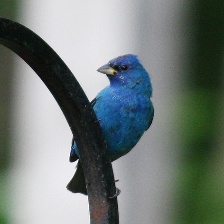} &
    \includegraphics[width=0.13\linewidth]{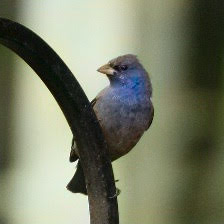} &
        \includegraphics[width=0.13\linewidth]{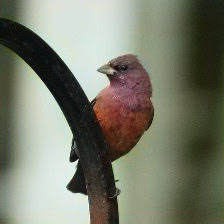} &
   \includegraphics[width=0.13\linewidth]{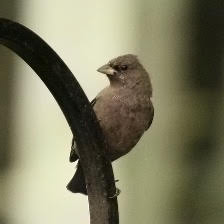} &
      \includegraphics[width=0.13\linewidth]{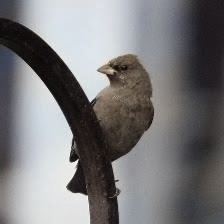}    & 
     \includegraphics[width=0.13\linewidth]{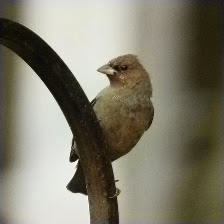} &
 \includegraphics[width=0.13\linewidth]{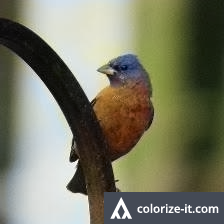}  \\

    \includegraphics[width=0.13\linewidth]{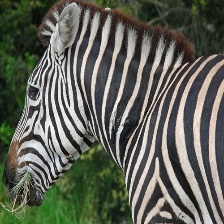} &
    \includegraphics[width=0.13\linewidth]{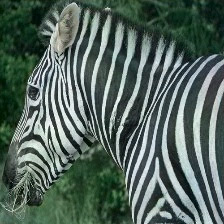} &
        \includegraphics[width=0.13\linewidth]{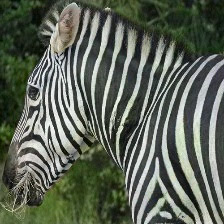} &
   \includegraphics[width=0.13\linewidth]{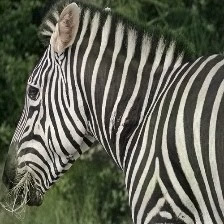} &
      \includegraphics[width=0.13\linewidth]{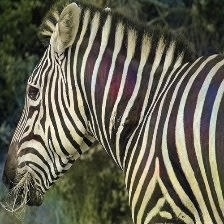}    &
     \includegraphics[width=0.13\linewidth]{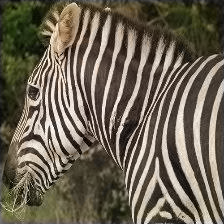} &
 \includegraphics[width=0.13\linewidth]{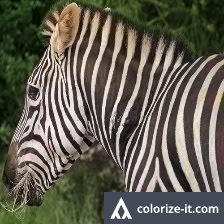}

        \end{tabular}
        \endgroup
\caption{Some qualitative results using, from right to left: Gray scale, ChromaGAN, ChromaGAN w/o Classification, ChromaNet, Iizuka \etal \cite{iizuka2016let}, Larsson \etal \cite{larsson2016learning} and Zhang \etal \cite{zhang2016colorful}}
\label{fig:ComparativeImages}
\end{figure*}

\section{Experimental Results and Discussion}\label{sec:results}
In this section the proposed method is evaluated both quantitatively and qualitatively. Notice that evaluating the quality of a colorized image in a quantitative way is a challenging task. 
For some objects several colors could perfectly fit. 
Therefore, quantitative measures reflecting how close the outputs are to the ground truth data are not the best measures for this type of problem. For that reason, qualitative comparisons are provided as well as a user-based perceptual study quantifying the realism of colorized images by the proposed method. Nevertheless, in order to quantitatively compare the results of the proposed methods to others in the literature a quantitative measure will also be used.

To assess the effect of each term of our loss function in the entire network, we perform an ablation study by evaluating three variants of our method: \emph{ChromaGAN}, using the adversarial and classification approach, \emph{ChromaGAN w/o Class} avoiding the classification approach  ($\lambda_s=0$) and \emph{ChromaNet} avoiding the adversarial approach ($\lambda_p=0$).


\subsection{Dataset}
We train each variant of the network end-to-end on 1.3M images from the subset of images~\cite{ILSVRC15} taken from ImageNet~\cite{imagenet_cvpr09}. It contains objects from 1000 different categories and of different color conditions, including grayscale images.
Due to the presence of fully connected layers in our network, the input size to the class distribution branch has to be fixed. We chose to work with input images of $224\times 224$ pixels as is done when training the VGG-16~\cite{simonyan2014very} on ImageNet. 
Thus, we have resized each image in the training set and converted it to a three channels grayscale image by triplicating the luminance channel $L$.

\subsection{Implementation Details}
We train the network for a total of five epochs and set the batch size to 10, on the 1.3M images from the ImageNet training dataset 
resized to $224 \times 224$. 
A single epoch takes approximately 23 hours on a NVIDIA Quadro P6000 GPU. The prediction of the colorization of a single image takes an average of 4.4 milliseconds. We minimize our objective loss using Adam optimizer with learning rate equal to $2e-5$ and momentum parameters $\beta_1=0.5$ and $\beta_2=0.999$. 
We alternate the optimization of the generator ${\cal{G}}_\theta$  and discriminator $D_w$. The first stage of the network (displayed in yellow in Fig. \ref{fig:model}), takes as input a grayscale image of size $224 \times 224$, and is initialized using the pre-trained weights of the VGG-16 \cite{simonyan2014very} trained on ImageNet.

\subsection{Quantitative Evaluation }

A perceptual realism study is performed to show the strength of coupling the adversarial approach with semantic information.
Furthermore, as state-of-the-art methods on colorization, a quantitative assessment of our method is included in terms of \emph{peak signal to noise ratio} (PSNR). 

\begin{table}[ht]
\begin{center}
\begin{tabular}{|l|c|}
\hline
 Method &  Naturalness \\
\hline\hline
 {Real images} & 87.1\% \\
 {ChromaGAN } & 76.9\% \\
 {ChromaGAN w/o Class} & 70.9\% \\
 {ChromaNet} & 61.4\% \\
   \hline
 {Iizuka \etal ~\cite{iizuka2016let}} & 53.9\%\\
Larsson \etal~\cite{larsson2016learning}  &  53.6\%  \\ 
Zhang \etal~\cite{zhang2016colorful}&  52.2\%    \\ 
   Isola \etal~\cite{isola2017image}& 27.6\%  \\
\hline
\end{tabular}
\end{center}
\caption{Numerical detail of the perceptual test. 
The values shows the mean naturalness over all the experiments of each method. }
\label{tab:perceptual-test}
\end{table}

The following perceptual realism study was performed. Images are shown one by one to non-expert participants, where some are natural color images and others the result of a colorization method such as ChromaGAN, ChromaGAN w/o classification, ChromaNet and {the four state-of-the-art methods \cite{iizuka2016let,larsson2016learning,zhang2016colorful,isola2017image}}.
For each image shown, the participant indicates if the colorization is realistic or not. Fifty images are taken randomly from a set of $1600$ images composed of $200$ ground truth images (from ImageNet and Places datasets~\cite{zhou2018places}) and $200$ results from each method (ChromaGAN, ChromaNet, ChromaGAN w/o Class and \cite{iizuka2016let,larsson2016learning,zhang2016colorful,isola2017image}).
The study was performed $62$ times. 
Table~\ref{tab:perceptual-test} shows the results of perceptual realism for each method. 
One can observe that ChromaGAN who couples the adversarial approach with semantic information yields perceptually more realistic results. Moreover, by comparing the results of ChromaNet, ChromaGAN w/o Class and~{\cite{iizuka2016let,larsson2016learning,zhang2016colorful,isola2017image}}, one can conclude that the adversarial approach plays a more important role than the semantic information for the generation of natural color images. 

Additionally, the PSNR of the obtained $(a,b)$ images is computed with respect to the ground truth and compared to those obtained for other fully automatic methods as shown in Table~\ref{tab:psnrTable}. The table shows the average of this measure over all the test images. One can observe that, in general, our PSNR values are higher than those obtained by~\cite{iizuka2016let,larsson2016learning,zhang2016colorful}. Moreover, comparing the PSNR of our three variants the highest one is achieved by ChromaNet. This is not surprising since the training loss of this method gives more importance to the quadratic color error term compared to the losses of ChromaGAN and ChromaGAN w/o Class. 

\begin{table}[t]
\begin{center}
\begin{tabular}{|l|c|}
\hline
 Method &  PSNR (dB) \\ 
\hline\hline
 {ChromaGAN} 
 & 24.98
 \\ 
 {ChromaGAN w/o Class}   
 & 25.04 \\
  {ChromaNet} 
  & \textbf{25.57} \\
  \hline
  Iizuka \etal~\cite{iizuka2016let}&  23.69   \\
  
  Larsson \etal~\cite{larsson2016learning}
& 24.93  \\ 
Zhang \etal~\cite{zhang2016colorful}&   22.04   \\ 
Isola \etal\cite{isola2017image} & 21.57\\
 
\hline
\end{tabular}
\end{center}
\caption{Comparison of the average PSNR values for automatic methods, some extracted from 
the table in~\cite{zhang2017real}. The experiment is performed on 1000 images of the ILSVRC2012 challenge set~\cite{russakovsky2015imagenet}.}
\label{tab:psnrTable}
\end{table}

\begin{figure*}[t]
\begin{center}
              
 \includegraphics[width=0.15\linewidth]{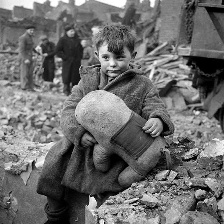}
      \includegraphics[width=0.15\linewidth]{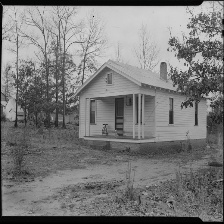}
     \includegraphics[width=0.15\linewidth]{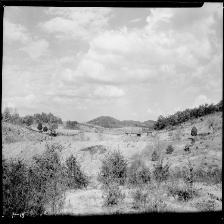}
    \includegraphics[width=0.15\linewidth]{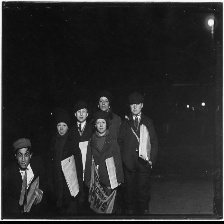}
         \includegraphics[width=0.15\linewidth]{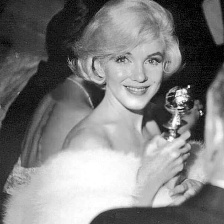}
         \includegraphics[width=0.15\linewidth]{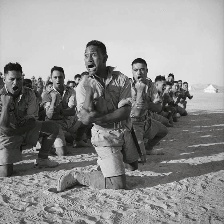}
    
 \includegraphics[width=0.15\linewidth]{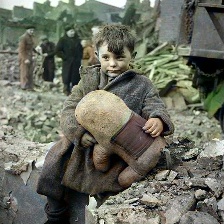}
 \includegraphics[width=0.15\linewidth]{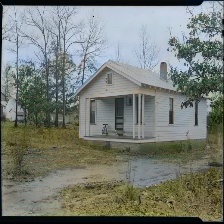}
 \includegraphics[width=0.15\linewidth]{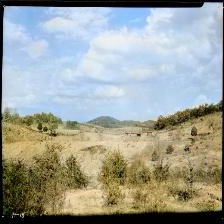}
 \includegraphics[width=0.15\linewidth]{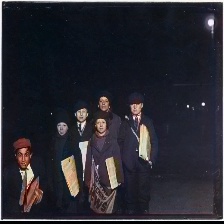} \includegraphics[width=0.15\linewidth]{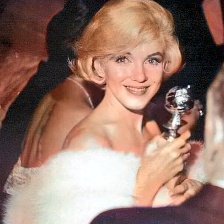}
          \includegraphics[width=0.15\linewidth]{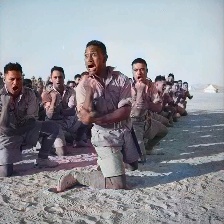}
\end{center}
   \caption{Colorization results of historical black and white photographs using the proposed ChromaGAN. Note that old black and white photographs are statistically different than actual ones, thus, making the process of colorize more difficult.}
\label{fig:historicalResults}
\end{figure*}

\begin{figure}[]
\begin{center}
\setlength{\tabcolsep}{1pt} 
            \begin{tabular}{ccc}  
            \textbf{GT} &  \textbf{Iizuka} \cite{iizuka2016let} & \textbf{ChromaGAN}\\
  \includegraphics[width=0.3\linewidth]{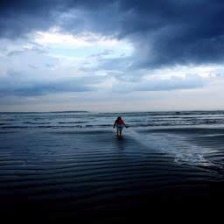} &
   \includegraphics[width=0.3\linewidth]{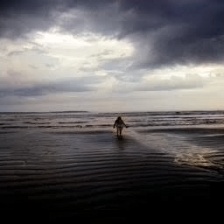} &
    \includegraphics[width=0.3\linewidth]{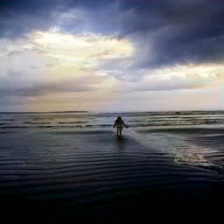} \\
 
   \includegraphics[width=0.3\linewidth]{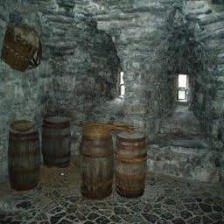} &
   \includegraphics[width=0.3\linewidth]{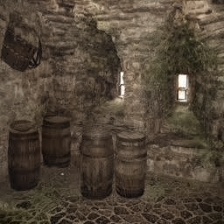} &
    \includegraphics[width=0.3\linewidth]{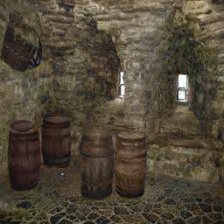} \\
    
      \includegraphics[width=0.3\linewidth]{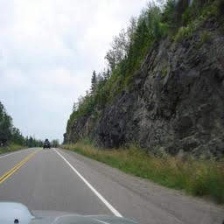} &
   \includegraphics[width=0.3\linewidth]{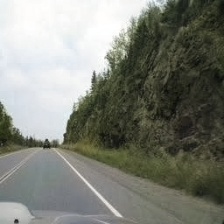} &
    \includegraphics[width=0.3\linewidth]{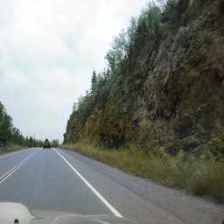}
    \end{tabular}
\end{center}
   \caption{Results on images from the validation set of the 
   Places Dataset. Left: Ground truth, middle: Iizuka \etal~\cite{iizuka2016let}, right: ChromaGAN. Notice that the model of~\cite{iizuka2016let} is trained on Places Dataset having the double of training images. Our model is trained on the ImageNet dataset. The results are comparable.}
\label{fig:Places}
\end{figure}

\begin{figure}[]
\begin{center}
\setlength{\tabcolsep}{1pt} 
            \begin{tabular}{ccc}  
            \textbf{GT}  & \textbf{Isola \cite{isola2017image}} &   \textbf{ChromaGAN} \\
  \includegraphics[width=0.3\linewidth]{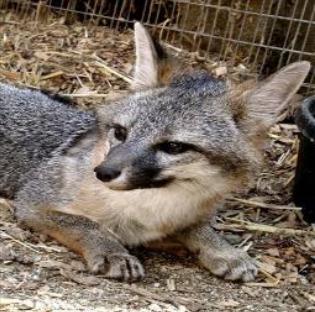} &
      \includegraphics[width=0.3\linewidth]{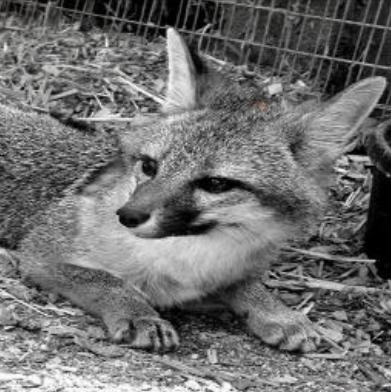} &
   \includegraphics[width=0.3\linewidth]{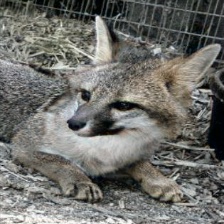} \\
 
  \includegraphics[width=0.3\linewidth]{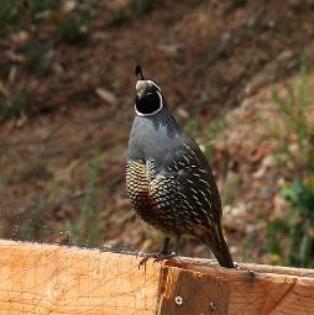} &
    \includegraphics[width=0.3\linewidth]{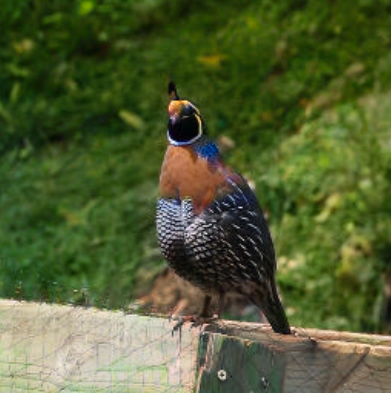}  &
   \includegraphics[width=0.3\linewidth]{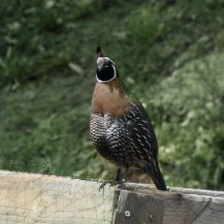} \\
 
   \includegraphics[width=0.3\linewidth]{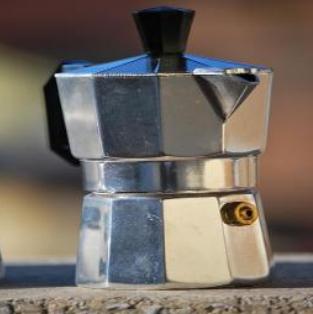}  &
    \includegraphics[width=0.3\linewidth]{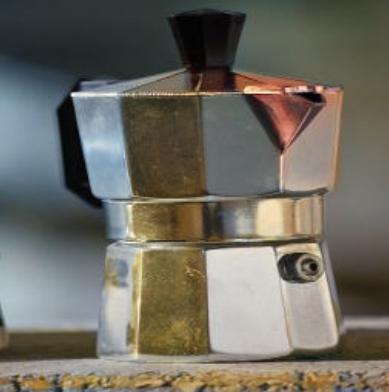} &
   \includegraphics[width=0.3\linewidth]{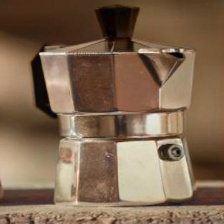}\\

    \end{tabular}
\end{center}
   \caption{Results on GAN-based approaches. From left to right: Ground Truth, results from Isola \cite{isola2017image} and results from ChromaGAN. The results from Isola \etal  were extracted from \cite{isola2017image}.}
\label{fig:isola-results}
\end{figure}

\subsection{Qualitative Evaluation}
Our results are compared to those obtained in~\cite{iizuka2016let, larsson2016learning, zhang2016colorful} by using the publicly available online demos. The methods are trained on ImageNet dataset in the case of~\cite{zhang2016colorful,larsson2016learning} and on Places dataset in the case of~\cite{iizuka2016let}. Notice that while ImageNet contains 1.3M training images, Places contains 2.4M. Several colorization results are shown on the validation set of ImageNet 
dataset in Fig.~\ref{fig:ComparativeImages} and on Places in Fig.~\ref{fig:Places}. As can be observed, the method in~\cite{iizuka2016let} and ChromaNet tend to output muted colours in comparison to the lively colors obtained with ChromaGAN, ChromaGAN w/o class and \cite{larsson2016learning, zhang2016colorful}. Also, ChromaGAN is able to reconstruct color information by adding natural and vivid colors in almost all the examples (specially the first, fifth, sixth, eighth and ninth rows).
Notice that the deep CNN in~\cite{zhang2016colorful} is trained with a multinomial cross entropy loss with re-balanced rare classes allowing in particular for unusual colors to appear in the colorized image. Desaturated results are mainly obtained by~\cite{iizuka2016let} and with our method without using the adversarial approach (specially in the first, second, third, fourth, fifth and eighth rows), in some cases also by~\cite{larsson2016learning} (second, fourth and eighth rows). Also, color boundaries are not clearly separated generally in the case of~\cite{iizuka2016let} and sometimes by GAN w/o Class (sixth row) and~\cite{larsson2016learning} (third, fourth and eighth rows). 
Inconsistent chromaticities can be found in the second and sixth row by~\cite{zhang2016colorful} where the wall is blue and the apples green and red at the same time. Third and seventh rows display some failure cases of our method: the bottom-right butterfly wing is colored in green. In fact, the case of the seventh row shows a difficult case for all the methods. 
For the sake of comparison, some results on Places dataset are shown in Fig.~\ref{fig:Places} for ChromaGAN trained on ImageNet, together with the results of~\cite{iizuka2016let} trained on Places dataset. In Fig.~\ref{fig:isola-results} the results of ChromaGAN are compared to those of the adversarial approach in~\cite{isola2017image}. The results in the second column of Fig.~\ref{fig:isola-results} were taken directly from~\cite{isola2017image}. Overall, their results provide sometimes more vivid colors as in the second row and sometimes uncolorized results as in the first row. Furthermore, the results of ChromaGAN yield better results in terms of consistent hue. In the second column unnatural color stains are observed as for instance the green marks under the bird (second row) and the yellow and red stains on the coffee pot (third row). 

\textbf{Legacy Black and White Photographs.} ChromaGAN is trained using color images where the chrominance information is removed. Due to the progress in the field of photography, there is a great difference in quality between old black and white images and modern color images. Thus, generating color information in original black and white images is a challenging task. Fig.~\ref{fig:historicalResults} shows some results. Additional examples can be found in the supplementary material, where we also include results applied on paintings.

\section{Conclusions}\label{sec:conclusion}
A novel colorization method is presented. The proposed ChromaGAN model is based on an adversarial strategy that captures geometric, perceptual and semantic information. A variant of ChromaGAN which avoids using the distribution of semantic classes also shows satisfying results. Both cases prove that our adversarial technique provides photo-realistic colorful images.
The quantitative comparison to state-of-the-art methods show that our method outperforms them in terms of PSNR and perceptual quality while qualitative comparisons show their visual high quality.

\section*{Acknowledgements}
This work was partially funded by 
MICINN/FEDER UE project, reference PGC2018-098625-B-I00 VAGS, and by H2020-MSCA-RISE-2017 project, reference 777826 NoMADS. We thank the support of NVIDIA Corporation for the donation of GPUs used in this work, Rafael Grompone von Gioi for valuable suggestions and Jos\'e Lezama for his help with the user study.

{\small
\bibliographystyle{ieee}
\bibliography{egbib}
}

\end{document}